\documentclass[nohyperref]{article}

\usepackage{microtype}
\usepackage{graphicx}
\usepackage{subfigure}
\usepackage{booktabs} % for professional tables

\usepackage{hyperref}

\usepackage[accepted]{icml2022}

\usepackage{amsmath}
\usepackage{amssymb}
\usepackage{mathtools}
\usepackage{amsthm}

\usepackage{amsmath}
\usepackage{bbm}
\usepackage{color, soul}
\usepackage{graphicx}
\usepackage{multirow}
\usepackage{caption}
\usepackage{wrapfig}
\usepackage{bm}
\usepackage{dsfont}
\usepackage{titlesec}

\usepackage[capitalize,noabbrev]{cleveref}

\theoremstyle{plain}

\theoremstyle{definition}

\theoremstyle{remark}

\usepackage[textsize=tiny]{todonotes}

\DeclareMathOperator*{\argmin}{arg\,min}

\titlespacing{\section}{0pt}{1pt}{1pt}
\titlespacing{\subsection}{0pt}{5pt}{1pt}
\icmltitlerunning{FORML: Learning to Reweight Data for Fairness}

\begin{document}

\setlength{\intextsep}{0.0pt}
\setlength{\abovecaptionskip}{0.1pt}
\setlength{\belowcaptionskip}{0.1pt}
\setlength{\abovedisplayskip}{0.1pt}
\setlength{\belowdisplayskip}{0.1pt}
\setlength{\textfloatsep}{0.0pt}

\twocolumn[
\icmltitle{FORML: Learning to Reweight Data for Fairness}



\icmlsetsymbol{equal}{*}

\begin{icmlauthorlist}
\icmlauthor{Bobby Yan}{equal,stanford}
\icmlauthor{Skyler Seto}{equal,apple}
\icmlauthor{Nicholas Apostoloff}{apple}
\end{icmlauthorlist}

\icmlaffiliation{stanford}{Department of Computer Science, Stanford University}
\icmlaffiliation{apple}{Apple}

\icmlcorrespondingauthor{Skyler Seto}{sseto@apple.com}

\icmlkeywords{Fairness, Data Reweighting, Meta-learning}

\vskip 0.3in
]



\printAffiliationsAndNotice{\icmlEqualContribution} % otherwise use the standard text.

\begin{abstract}
    Machine learning models are trained to minimize the mean loss for a single metric, and thus typically do not consider fairness and robustness. Neglecting such metrics in training can make these models prone to fairness violations when training data are imbalanced or test distributions differ.  This work introduces \textbf{F}airness \textbf{O}ptimized \textbf{R}eweighting via \textbf{M}eta-\textbf{L}earning (FORML), a training algorithm that balances fairness and robustness with accuracy by jointly learning training sample weights and neural network parameters. The approach increases model fairness by learning to balance the contributions from both over- and under-represented sub-groups through dynamic reweighting of the data learned from a user-specified held-out set representative of the distribution under which fairness is desired. FORML improves  equality of opportunity fairness criteria on image classification tasks, reduces bias of corrupted labels, and facilitates building more fair datasets via data condensation. These improvements are achieved without pre-processing data or post-processing model outputs, without learning an additional weighting function, without changing model architecture, and while maintaining accuracy on the original predictive metric.
\end{abstract}

\section{Introduction}
\label{intro}
Deep neural networks (DNNs) are widely used for machine learning applications, including image classification \citep{krizhevsky2012imagenet}, speech recognition \citep{hinton2012deep}, natural language understanding \citep{devlin2018bert}, and healthcare \citep{esteva2019guide}.  Despite the strong predictive performance of modern DNN architectures, when the distribution of the evaluation data differs from that of the training data, or the test evaluation metrics differ from those during training, models can inherit biases and fail to generalize due to spurious correlations in the dataset and overfitting to the training metric. Importantly, this can result in fairness violations for certain groups in the test set \citep{hardt2016equality}.  This issue is  exacerbated as notions of fairness and accuracy may be inherently opposed to one another. % In particular, it has been stated that demanding fairness of models comes at the cost of reduced predictive accuracy \citep{berk2017convex,kearns2019ethical}.

A common data-centric paradigm in fairness and robustness for mitigating data distribution shift and class imbalance is through data reweighting. Classical approaches to data reweighting involve resampling data \citep{kahn1953methods}, using domain-specific knowledge \citep{zadrozny2004learning}, estimating weights based on data difficulty \citep{lin2017focal, malisiewicz2011ensemble}, and using class-count information \citep{cui2019class, roh2021sample}. 

Data-dependent reweighting methods compute weights by reweighting iteratively based on the training loss \citep{fan2018learning, petrovic2020fair, wu2018learning}, or on fairness violations on the training set \citep{jiang2020identifying}.  Other works have extended classical approaches by learning a (typically parameterized) function mapping the inputs to weights \citep{zhao2019metric,lahoti2020fairness}, or by   optimizing for the weights by treating them as directly learnable parameters  \citep{ren2018learning}, however their approach does not learn a global set of weights and cannot be used for post-training data compression.  Prior reweighting approaches  aim to  improve generalization and robustness to noisy labels \citep{ren2018learning, saxena2019data, shu2019meta, vyas2020learning},  class-imbalance \citep{lin2017focal, kumar2010self, dong2017class}, or  training time and convergence by learning a curriculum over instances \citep{kumar2010self, saxena2019data}. Few of these works aim to reweight data to directly optimize an additional (fairness) metric \citep{jiang2018mentornet}, \citep{zhao2019metric}.

\begin{figure}
    \centering
    \includegraphics[width=2in]{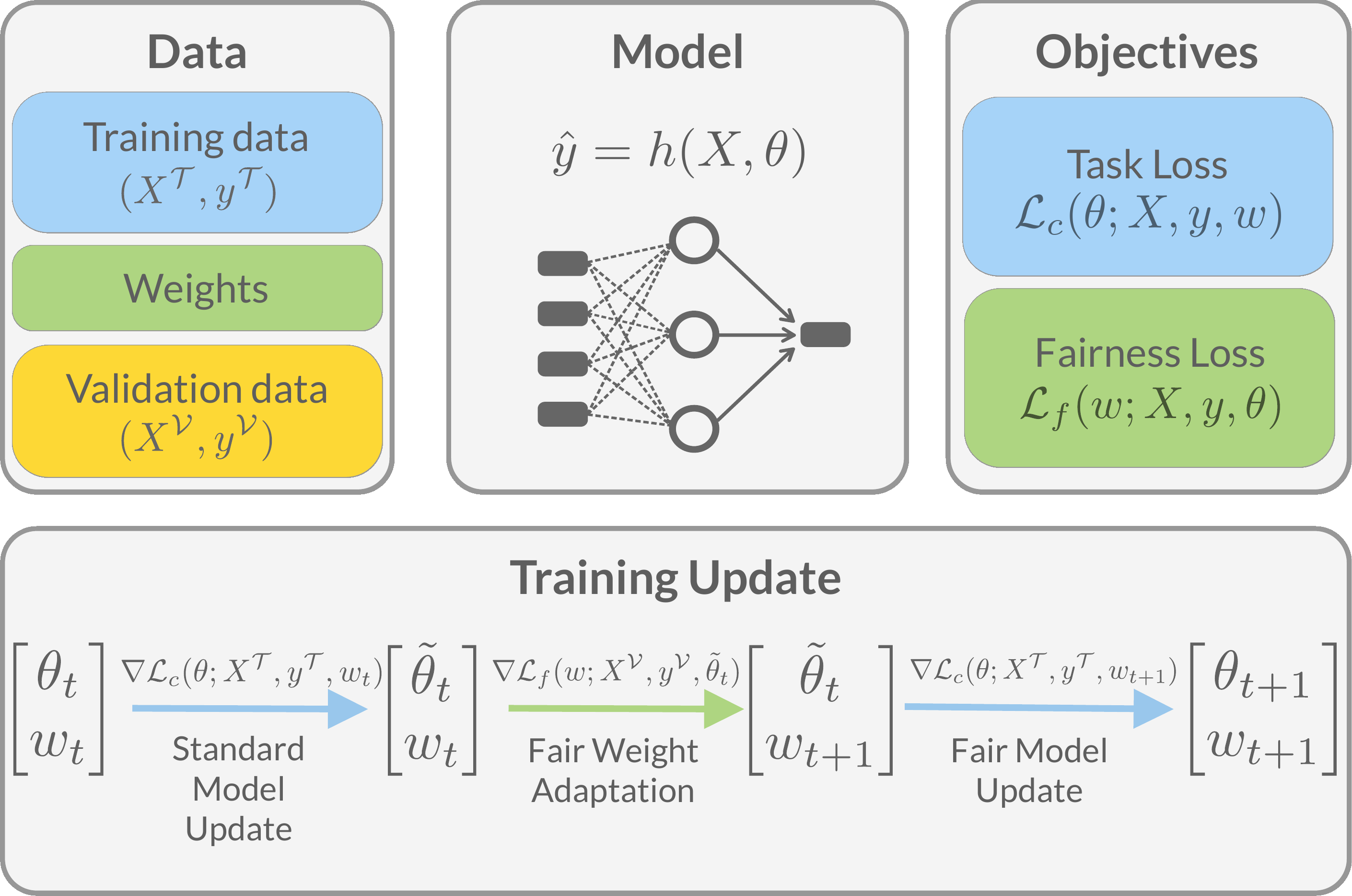}
    \caption{An overview of the FORML algorithm, which optimizes sample weights $w$ and model parameters $\theta$ using the meta-learning paradigm.}
    \label{fig:overview}
\end{figure}

In this work, we propose \textbf{F}airness \textbf{O}ptimized \textbf{R}eweighting via \textbf{M}eta-\textbf{L}earning (FORML), an algorithm that directly optimizes both fairness and predictive performance.  We follow the learning-to-learn (meta-learning) paradigm \citep{andrychowicz2016learning, finn2017model, lake2015human, ren2018learning, thrun2012learning} and jointly learn a weight for each sample in the training set and the model parameters by optimizing for the given test metric and a fairness criteria over a held-out exemplar set, which incorporates data importance into optimization.  At a high-level, our algorithm (1) optimizes model parameters via a weighted loss objective, and (2) optimizes a global set of sample weights using a given fairness metric over the exemplar set. An overview of the gradient rules of the algorithm are given in \autoref{fig:overview}. Learning sample weights over an exemplar set helps adapt the model to the fairness metric similar to how the model may be evaluated at test time, allows any practitioner to define fairness  non-parametrically, and improves label efficiency as held-out validation sets are typically much smaller than the training data set where attribute labels are needed.% This leads the model to better balance the performance of the different metric across sub-groups which can lead to improvements in the discrepancy between train and test performance when the sub-group distributions differ.   % Our algorithm resembles the model-agnostic meta-learning (MAML) framework \citep{finn2017model, ren2018learning} as FORML learns model parameters by optimizing the original training metric $\mathcal{L}_c$ and performs a gradient update over the validation set with a meta-objective (fairness metric $\mathcal{L}_f$) to learn sample weights, and finally adapts the model parameters with the new fairly reweighted objective.

We experiment with  image recognition datasets and demonstrate that our approach reduces fairness violations (measured by equality of opportunity) by improving worst group performance without harming overall performance. FORML does not simply learn to reweight based on the number of samples as we see reduced fairness violations even when samples are uniformly distributed across groups, but on the importance of data points to the fairness criteria. Further, FORML improves performance in noisy label settings, and FORML can be used to remove harmful data samples leading to improved fairness and higher data efficiency.

% FORML extends existing meta-learning and fairness reweighting approaches including \citep{finn2017model, jiang2020identifying, ren2018learning, zhao2019metric} as (1) the learned weights are global for each datum rather than local to a batch and can be leveraged for insights into detecting bias, (2) weights are directly optimized based on a fairness metric evaluated on a with-held validation set to better model test-time model evaluation, and (3) the approach does not require learning a parameterized weighting function avoiding specification of a complex weighting function and placing the focus of training on the data. 

\section{Learning to Reweight Data for Optimizing Fairness}

% \textsc{FORML} is inspired by prior modeling and optimization techniques, most notably example reweighting \citep{ren2018learning} and fairness constrained optimization \citep{cotter2019optimization, jiang2020identifying}.

Consider a dataset $\mathcal D = \{(x^{(i)}, y^{(i)}, a^{(i)}) \mid 1 \leq i \leq N\}$, where $x^{(i)} \in \mathcal X$ is the input (e.g., images, texts), $y^{(i)} \in \mathcal Y$ is the target variable, and $a^{(i)} \in \mathcal A$ is the sensitive attribute.  Let $D_{\mathcal{T}}$ and $D_{\mathcal{V}}$ denote the train and exemplar sets formed from $\mathcal{D}$, and $(x_{\mathcal{T}}^{(i)}, y_{\mathcal{T}}^{(i)}, a_{\mathcal{T}}^{(i)})$, $(x_{\mathcal{V}}^{(i)}, y_{\mathcal{V}}^{(i)}, a_{\mathcal{V}}^{(i)})$ be samples from $D_{\mathcal{T}}$ and $D_{\mathcal{V}}$ respectively. Let $h(x; \theta) \in \mathcal H$ be a classifier parameterized by model parameters $\theta$. The main objective is to learn parameters of $h$ that minimize the expected loss: $\theta^* = \underset{\theta}{\argmin} \underset{(x^{(i)}, y^{(i)}, a^{(i)}) \sim D_{\mathcal{T}}}{\sum} \mathcal{L}_c(h(x^{(i)}; \theta), y^{(i)}).$

Improving fairness with minimal impact on accuracy can be formulated as a constrained optimization problem that depends on fairness violation thresholds $\epsilon_i$:

\vspace{-6mm}
\begin{equation}
  \begin{aligned}
    \theta^*  =
      & \argmin_{\theta}  \sum_{(x^{(i)}, y^{(i)}, a^{(i)}) \sim D_{\mathcal{T}}} \mathcal{L}_c(h(x^{(i)}; \theta), y^{(i)})\\
      & \ s.t. \sum_{(x^{(i)}, y^{(i)}, a^{(i)}) \sim D_{\mathcal{T}}} \mathcal{L}_f(h(x^{(i)}; \theta), y^{(i)};  a^{(i)}) \leq \epsilon_i.
  \end{aligned}
   \label{eq:const_opt}
\end{equation}
\vspace{-6mm}
 % \theta^* = \argmin_{\theta} \mathbb E_{(x_i, y_i) \sim D} L(h(x_i; \theta), y_i),

Classical algorithms for solving the penalized objective are based on the basic differential multiplier method \citep{platt1987constrained}, which defines a gradient update rule for both model parameters and the weight parameters.  Gradient-based approaches to hyper-parameter optimization have also been proposed in \citep{bengio2000gradient}.

 Constrained optimization with Lagrangians have been studied more recently for fairness \citep{cotter2019optimization, cotter2019two}.  Recent work studied \autoref{eq:const_opt} in the context of label bias and proposed an iterative reweighting algorithm based on the fairness violation on the training set \citep{jiang2020identifying}.  Other work demonstrated that optimizing model parameters over a given evaluation metric is equivalent to optimizing over an example weighted loss \citep{zhao2019metric}. 

An alternative view for constrained optimization of \citep{jiang2020identifying} is that the model is adapted by the training sample weights (learned through the constraint) to handle new tasks where the constraint is enforced.  Such an approach is typically handled well by meta-learning optimization algorithms \citep{fallah2021generalization, finn2017model}.  We follow a similar procedure to MAML adaptation algorithms \citep{finn2017model, ren2018learning} and consider a two-step procedure similar to that of \citep{jiang2020identifying}.  We propose an iterative optimization procedure, where the model parameters $\theta_{t+1}$ and sample weights $w_{t+1}$ at the $t$th timestep are jointly optimized according to the learned weighted training objectives:
\begin{align}
   w_{t+1} &= \argmin_{w} \mathbb E_{(x, y, a) \sim D_{\mathcal{V}}} \left[ \mathcal{L}_f(h(x; \tilde{\theta}), y; a) \right],\\
   \theta_{t+1} &= \argmin_{\theta}\mathbb E_{(x, y) \sim D_{\mathcal{T}}} \left\langle w_{t+1}, \mathcal{L}_c(h(x; \theta), y)\right\rangle,
\end{align}

where $\tilde{\theta} = \theta_t - \eta_c \nabla_{\theta} \mathcal{L}(\theta) \Bigr|_{\substack{\theta=\theta_t}}$ is the one gradient descent step update over the training loss.  The approach can be summarized as follows: (1) model parameter updates are computed based on the training loss and the model parameters are evaluated on the fair loss on the exemplar set to determine weights that offset the impact to fairness, and (2) model parameters are adapted according to the weight parameters such that the weight parameters do not negatively impact fairness. 

% The proposed procedure is named Fairness Optimized Reweighting via Meta-Learning (FORML) and can be distinguished from prior approaches as the sample weights can be efficiently optimized via meta-learning by taking a gradient step over a clean held-out set with the fairness metric (meta-objective).  An overview of the meta-learning optimization procedure is given in \autoref{fig:overview_pass}.   % First, the model performs a gradient step on the predictive loss. Second, the weight parameters are updated through evaluation over the validation set.  Finally, the model parameters undergo a meta-update according to the training sample weights learned over the validation set effectively adapting the loss that balances fairness and predictive performance similar to the algorithm of \citep{ren2018learning}.

Advantages of the proposed meta-learning procedure over a fixed update rule such as \citep{jiang2020identifying} and prior meta-learning reweighting methods including \citep{finn2017model, ren2018learning, shu2019meta} are that FORML (1) incorporates gradient information from the fairness metric to the gradient updates of the model parameters directly adapting the model parameters for fairness, (2) adapts sample weights over exemplar data to better generalize to fairness on the test set by optimizing  both (user-defined exemplar) fairness and performance, and (3) learns global weights over training which can be used to learn data importance and difficulty over weights local to a batch. 

\subsection{Implementation of \textsc{FORML}}

\begin{figure}
    \centering
    \includegraphics[width=3in]{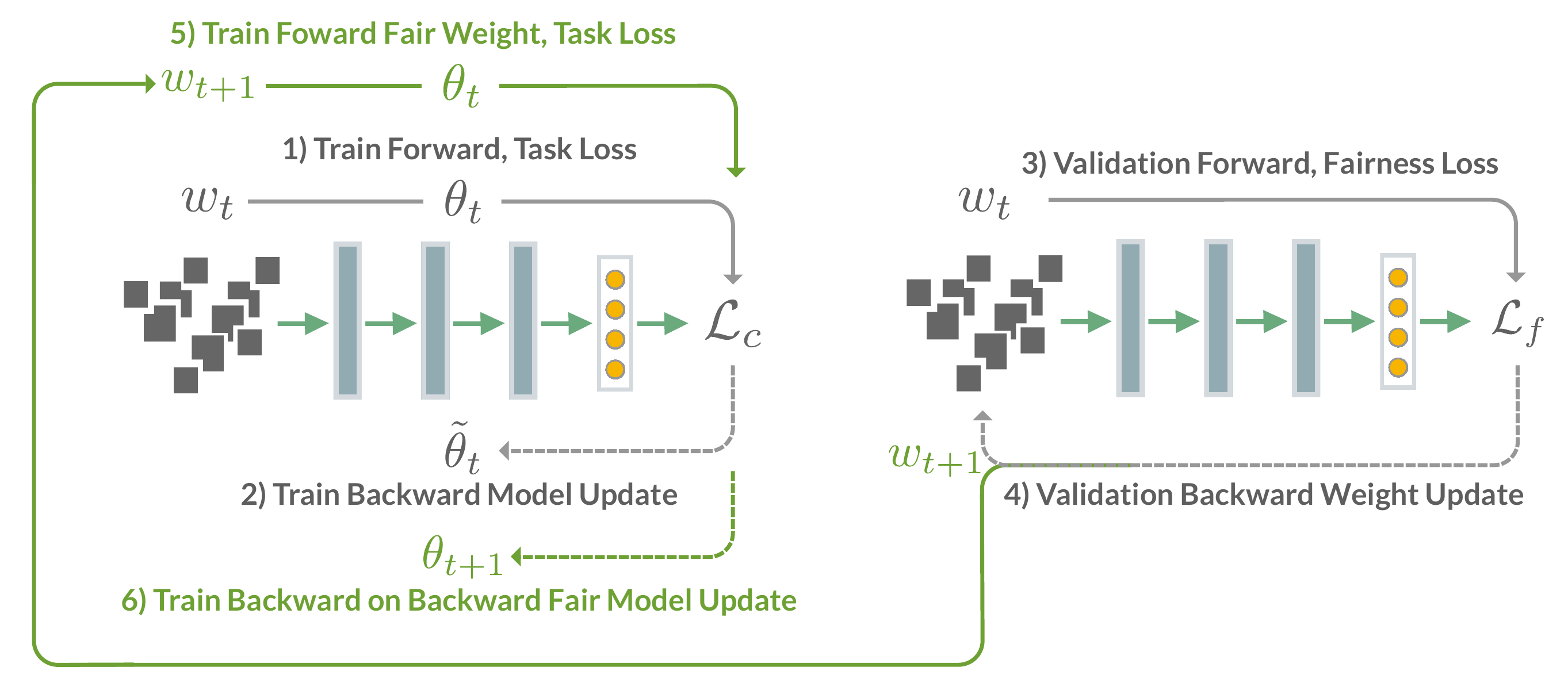}
    \caption{Illustration of the forward and backward passes for the model parameters and training sample weights.}
    \label{fig:overview_pass}
\end{figure}

Training a model with FORML involves two additional gradient updates over the standard stochastic optimization pipeline. Pseudocode for computation of FORML is given in \Cref{algorithm:forml} and an overview in \autoref{fig:overview_pass}. 

The algorithm is as follows: (1) sample the training set uniformly and exemplar set to ensure a small pre-specified number of samples per class for fairness evaluation, (2) update training example weights by first computing a model update using SGD (Steps 6-8) then updating the example weights based on the gradient\footnote{Note that in \Cref{algorithm:forml}, we perform only one gradient update for the computation of the updated model parameters and training sample weights for efficiency, however using multiple gradient updates or running to convergence is a straightforward extension.} of the fairness metric on the exemplar dataset (Steps 9-11), and (3) update the model parameters using the updated sample weights over the original training metric (Steps 12-13) to avoid mixing gradient information from the previous sample weights, resulting in a model which has been adapted to handle both the training metric and fairness metric.  % For deep neural networks trained with gradient-based optimization algorithms, FORML's update rules require the computation of a gradient from a prior gradient update as in \citet{finn2017model}. These gradient updates can be easily implemented using automatic differentiation techniques, implemented in popular software packages such as PyTorch and TensorFlow \citep{pytorch, abadi2016tensorflow}.

\begin{algorithm}[ht]
  \linespread{1.25}\selectfont
  \caption{\textsc{FORML}: Fairness-Optimized Reweighting via Meta-Learning}
  \label{algorithm:forml}
\begin{algorithmic}[1]
   \STATE {\bfseries Input:} $\mathcal D_T$, $\mathcal D_V$,  $\theta_0$,  $w_0$,  $\mathcal{L}_c$,  $\mathcal{L}_f$,  $\eta_c$, $\eta_f$\\[3pt]
   \FOR{$t=0$ {\bfseries to} $T - 1$}
    \FOR{$b=0$ {\bfseries to} $B - 1$}
        \STATE $X_{\mathcal T}, y_{\mathcal T} \gets$ sample($D_{\mathcal T}$, $b$), 
        \STATE $X_{\mathcal V}, y_{\mathcal V} \gets$ sample\_fair($D_{\mathcal V}$, $b$)
        \STATE $\hat y_{\mathcal T} \gets h(X_{\mathcal T}; \theta_t)$\\[3pt]

        \STATE $\delta_c =  \sum_{i=1}^n \nabla_{\theta_t} \left[\frac{ exp(w_t^{(i)}) }{\sum_{j} { exp( w_t^{(j)}} ) } \mathcal{L}_c \left(y^{(i)}_{\mathcal T}, \hat y^{(i)}_{\mathcal T}\right)\right]$\\[3pt]

        \STATE $\tilde \theta_t \gets \theta_t - \eta_c \delta_c$
        \STATE $\hat y_{\mathcal V} \gets h(X_{\mathcal V}; \tilde \theta_t)$\\[3pt]

        \STATE $\delta_f =  \sum_{i=1}^n \nabla_{w_t} \left[\mathcal{L}_f\left(y^{(i)}_{\mathcal V}, \hat y^{(i)}_{\mathcal V}\right)\right]$\\[3pt]

        \STATE $w_{t+1} \gets w_t - \eta_f \delta_f$\\[3pt]
        \STATE ${\tilde \delta_c} =  \sum_{i=1}^n \nabla_{\theta_t} \left[  \frac{ exp(w_{t+1}^{(i)}) }{\sum_{j} { exp( w_{t+1}^{(j)}} ) } \mathcal{L}_c\left(y^{(i)}_{\mathcal T}, \hat y^{(i)}_{\mathcal T}\right)\right]$\\[3pt]

        \STATE $\theta_{t+1} \gets \theta_t - \eta_f {\tilde \delta_c}$
    \ENDFOR
   \ENDFOR
\end{algorithmic}
\end{algorithm}

\section{Image Classification Results}
\label{results}

We  experiments on image recognition datasets to demonstrate FORML mitigates bias in datasets. In all experiments,  FORML reduces fairness violations while maintaining accuracy\footnote{This is a relatively under-explored area of fairness which typically aims to minimize only fairness violations.}.  While other works further reduce fairness violations as in \citep{jiang2020identifying}, doing so typically results in a decrease in accuracy and is different from our aims.

\subsection{Experimental Setup}
We compare FORML with prior weighting schemes for improving fairness and robustness, including static weighting strategies: uniform, random, and class-balanced weighting \citep{cui2019class}, and learned weighting strategies: LB \citep{jiang2020identifying}, MOEW \citep{zhao2019metric}, and MWN \citep{shu2019meta}.  In our experiments, we take the exemplar set as the pre-defined validation set if the dataset has one, or create one by sampling uniformly (by attribute) from the training set.

\subsection{CIFAR Experiments}
\label{class_fairness}

In this section, we evaluate \textsc{FORML} using CIFAR-10 and CIFAR-100, and investigate whether FORML learns training sample weights that improve fairness. For these experiments, there is no pre-defined sensitive attribute, therefore we use the class label as the sensitive attribute, and the objective is to achieve equal classification performance across classes reported as TPRD.

\begin{table}[t]
  \centering
  \resizebox{\columnwidth}{!}{
  \begin{tabular}{lccc}
    \toprule
    Method                    & TPRD (\%)  $\downarrow$           & maxFNR (\%) $\downarrow$          & Accuracy (\%) $\uparrow$    \\
    \midrule
    Uniform / Proportion      & $8.79 \pm 0.14$       & $10.80 \pm 0.14$      & $\bm{95.25 \pm 0.03}$\\
    Random                    & $11.22 \pm 0.15$      & $13.84 \pm 0.85$      & $93.77 \pm 0.04$ \\
    MOEW                      & $23.08 \pm 0.26$      & $30.88 \pm 0.23$      & $83.072 \pm 0.03$ \\
    MWN & $9.0 \pm 0.15$ & $11.17 \pm 0.14$ & $95.10 \pm 0.03$\\
    LB                 &      $8.68 \pm 0.15$                 &   $10.73 \pm 0.14$                    &    $95.21 \pm 0.03$\\
    FORML                     & $\bm{7.97 \pm 0.12}$  & $\bm{10.36 \pm 0.17}$ & $95.20 \pm 0.03$ \\
    \bottomrule
  \end{tabular}
  }
    \caption{Fairness metrics: True Positive Rate Disparity (TPRD), Maximum False Negative Rate (maxFNR), and test accuracy for ResNet-18 on CIFAR-10. The means and standard errors are from 30 runs.}
    \label{cifar10-table}
\end{table}

We train a ResNet-18 architecture \citep{he2016deep} and summarize results in \autoref{cifar10-table}. Neither random nor MOEW are able to perform competitively with other methods, and FORML achieves $\sim 1\%$ improvement (a relative performance gain of $\sim 10\%$) over uniform and label bias on CIFAR-10. 

\begin{table}[t]
  \centering
  \resizebox{\columnwidth}{!}{
  \begin{tabular}{lccc}
    \toprule
    Method                    & TPRD (\%)  $\downarrow$           & maxFNR (\%) $\downarrow$          & Accuracy (\%) $\uparrow$    \\
    \midrule
    Uniform / Proportion  &  $46.27 \pm 0.46$      & $49.29 \pm 0.43$      & $\bm{77.80 \pm 0.06}$  \\
    Random            & $50.56 \pm 0.43$      & $55.56 \pm 0.38$      & $72.54 \pm 0.08$        \\
    MOEW          & $67.2 \pm 0.72$      &  $80.6 \pm 0.68$      & $53.94 \pm 0.03$             \\
    MWN & $45.83 \pm 0.46$ & $49.13 \pm 0.45$ & $77.62 \pm 0.05$ \\
    LB               &   $45.87 \pm 0.52$                   &   $49.17 \pm 0.46$                    &    $77.73 \pm 0.05$      \\
    FORML          & $\bm{45.17 \pm 0.35}$ & $\bm{48.80 \pm 0.32}$ & $77.51 \pm 0.08$           \\
    \bottomrule
  \end{tabular}
  }
    \caption{Performance with ResNet-18 on CIFAR-100. The means and standard errors are from 30 runs.}
    \label{cifar100-table}
\end{table}

For CIFAR-100, we see a similar trend in Table~\ref{cifar100-table}, where only FORML is able to outperform the uniform training method while simultaneously maintaining accuracy.  

\subsection{Celeb-A Experiments}
\label{sens_fairness}

We train a ResNet-18 architecture \citep{he2016deep} to predict the binary attribute ``Attractive''. The sensitive attribute is gender denoted by ``Male''. For Celeb-A, we use the pre-defined train, val, test split and compare models at the epoch with the highest validation accuracy during training. Metrics are reported over the test set in \autoref{celeba-table}. 
 
  \begin{table}[t]
  \centering
  \resizebox{\columnwidth}{!}{
  \begin{tabular}{lccc}
    \toprule
    Method                    & TPRD (\%)  $\downarrow$           & maxFNR (\%) $\downarrow$          & Accuracy (\%) $\uparrow$    \\
    \midrule
    % Uniform & $32.80 \pm 4.01$ & $59.08 \pm 7.71$ & $78.00 \pm 1.47$ \\
    % Random & & & & 
    Uniform  &$24.91 \pm 1.67$ & $36.51 \pm 1.78$ & $81.72 \pm 0.04$\\
    Random  & $24.52 \pm 1.40$ & $34.59 \pm 1.68$ & $80.63 \pm 0.14$\\
    Proportion & $23.44 \pm 1.71$ & $34.68 \pm 1.79$ & $81.71 \pm 0.10$\\
    LB  & $24.63 \pm 1.62$ & $36.94 \pm 2.63$ & $\bm{81.91 \pm 0.07}$\\
    % % celeba-epochs_10-sensitive_Male-target_Attractive-reweight_reverse-meta_loss_MeanLossD-lr_w_scale_1.0-numpc_val_batch_20-deterministic
    FORML & $\bm{22.32 \pm 1.7}$ & $\bm{31.14 \pm 1.87}$ & $81.63 \pm 0.24$\\     
    % FORML & $\bm{18.70 \pm 2.73}$ & $\bm{25.78 \pm 3.87}$ & $80.07 \pm 1.06$\\
    \bottomrule
  \end{tabular}
  }
    \caption{Celeb-A attribute prediction performance with ``Attractive'' target attribute and ``Gender'' protected attribute. The means and standard errors are from 5 runs.}
    \label{celeba-table}
\end{table}

Both proportional reweighting based on sensitive attribute and random weightings perform better than uniform weighting on maxFNR and proportional reweighting improves TPRD by 1\%, however the label bias approach offers no advantage over uniform training.   In contrast, FORML achieves a reduction of $\sim 2\%$ (a relative performance gain of $\sim 10\%$)  on the fairness violation (TPRD), and a reduction of $\sim 5\%$ in maxFNR (a relative performance gain of $\sim 15\%$), while retaining accuracy.  

\section{Data Robustness and Condensation}

\subsection{Mislabeled Data in MNIST}
\label{mislabel}

We investigate the performance of FORML with mislabeled data, where the objective is to learn data weights and ignore wrongly labeled data.  To simulate this setup, we take the MNIST dataset and randomly change $20\%$ of the training data labels to ``$2$'', as in \citep{jiang2020identifying}.  Based on the results in \autoref{mnist-corrupt-table}, FORML achieves higher accuracy and reduces fairness violation over training with uniform weighting and label bias weightings. In the supplementary material, we compared the ratio of weighted loss of samples labeled ``2'' to other samples demonstrating FORML learns less from mislabeled samples than uniform weighting. %take test accuracy at the highest validation accuracy\footnote{We do not use a validation set for the baseline and label bias as the baselines do not use one in \citet{jiang2020identifying} and reserving the validation set dropped overall test performance at the end of training on MNIST.}.  

\begin{table}[t]
  \centering
\resizebox{\columnwidth}{!}{
  \begin{tabular}{lccc}
    \toprule
    Method                    & TPRD (\%) $\downarrow$ & maxFNR (\%) $\downarrow$     &  Accuracy (\%) $\uparrow$    \\
    \midrule
    Uniform (clean data) & $5.21 \pm 0.40$ & $6.12 \pm 0.40$ & $97.16 \pm 0.12$\\
    \bottomrule
    Uniform              &  $8.59 \pm 0.91$ & $10.00 \pm 0.89$ & $95.24 \pm 0.36$\\
    LB  & $7.71 \pm 0.76$ & $9.05 \pm 0.76$ & $95.52 \pm 0.27$\\
    FORML                   & $\bm{6.93 \pm 0.79}$ & $\bm{8.22 \pm 0.83}$ & $\bm{95.90 \pm 0.21}$\\
    \bottomrule
  \end{tabular}
  }
    \caption{Performance of a two-hidden-layer MLP on the MNIST dataset with 20\% mislabeled data. The means and standard errors are reported over 10 runs.}
    \label{mnist-corrupt-table}
\end{table}
% \vspace{-\baselineskip}

\subsection{Using Data Weights to Build Fair Datasets}
\label{data_weights}

FORML improves fairness by identifying representative samples and reducing impact of harmful samples during training. We extend this analysis and investigate if FORML can post-hoc condense datasets yielding a more fair dataset.  Specifically, we train a model with FORML using an exemplar set of $1000$ samples with the lowest number of forgetting events \citep{toneva2018empirical}. We then remove 10\% of the training samples with the lowest normalized weighted loss, computed using an  exponential moving average (EMA) over training epochs. Using a moving average incorporates weight impact over all of training rather than only considering the weights at the end of training.

\begin{table}[ht]
  \centering
  \resizebox{\columnwidth}{!}{
  \begin{tabular}{lccc}
    \toprule
    Method          &  TPRD (\%)  $\downarrow$           & maxFNR (\%) $\downarrow$          & Accuracy (\%) $\uparrow$    \\
    \midrule
    Uniform 50k & $46.0 \pm 0.85$ & $49.6 \pm 0.92$ & $\bm{77.74 \pm 0.18}$ \\
    Random 45k & $46.8 \pm 0.33$  & $50.0 \pm 0.49$  & $76.7 \pm 0.09$ \\
    \midrule
    FORML 45k & $\bm{43.8 \pm 0.52}$ & $\bm{47.6 \pm 0.36}$ & $77.50 \pm 0.04$ \\
    \bottomrule
  \end{tabular}
  }
    \caption{Performance of ResNet-18 re-training on CIFAR-100 with data removal.  The means and standard errors are reported over 5 runs.}
    \label{condensation-table}
\end{table}

To demonstrate performance with the sub-sampled dataset, we re-train the model using standard training with the sub-sampled dataset; results are summarized in \autoref{condensation-table} and indicate that on CIFAR-100, removing samples based on the weighted EMA of the loss performs similarly and reduces fairness violation over training with the full dataset.  Training on the sub-sampled dataset also leads to better results than a model that has been trained on a random subset of the same size.  This indicates that FORML weights can build fairer datasets through removing harmful samples.

\section{Conclusion}
\label{conc}

 In this work, we present FORML, an algorithm for dynamically reweighting data to train fair networks, and identify and mitigate bias from erroneous or over-represented samples. FORML has several benefits: it is simple to implement by making small modifications in training, and requires neither data pre-processing nor post-processing of the model outputs.  Further, FORML is a data-centric method that improves fairness based on the dataset and is agnostic to the model and fairness metric, extending beyond classification. We believe this work is a step towards creating fairer models without sacrificing accuracy by better leveraging data.  

\section*{Acknowledgements}
We are grateful to Carlos Guestrin, Katherine Metcalf, Barry-John Theobald, and Russ Webb for their helpful discussions, and comments on the ideas in this work.

\bibliographystyle{plainnat}
\bibliography{reference}

\begin{thebibliography}{36}
\providecommand{\natexlab}[1]{#1}
\providecommand{\url}[1]{\texttt{#1}}
\expandafter\ifx\csname urlstyle\endcsname\relax
  \providecommand{\doi}[1]{doi: #1}\else
  \providecommand{\doi}{doi: \begingroup \urlstyle{rm}\Url}\fi

\bibitem[Andrychowicz et~al.(2016)Andrychowicz, Denil, Gomez, Hoffman, Pfau,
  Schaul, Shillingford, and De~Freitas]{andrychowicz2016learning}
Marcin Andrychowicz, Misha Denil, Sergio Gomez, Matthew~W Hoffman, David Pfau,
  Tom Schaul, Brendan Shillingford, and Nando De~Freitas.
\newblock Learning to learn by gradient descent by gradient descent.
\newblock In \emph{Advances in neural information processing systems}, pages
  3981--3989, 2016.

\bibitem[Bengio(2000)]{bengio2000gradient}
Yoshua Bengio.
\newblock Gradient-based optimization of hyperparameters.
\newblock \emph{Neural Comput.}, 12\penalty0 (8):\penalty0 1889--1900, 2000.

\bibitem[Cotter et~al.(2019{\natexlab{a}})Cotter, Jiang, Gupta, Wang, Narayan,
  You, and Sridharan]{cotter2019optimization}
Andrew Cotter, Heinrich Jiang, Maya~R Gupta, Serena Wang, Taman Narayan,
  Seungil You, and Karthik Sridharan.
\newblock Optimization with non-differentiable constraints with applications to
  fairness, recall, churn, and other goals.
\newblock \emph{J. Mach. Learn. Res.}, 20\penalty0 (172):\penalty0 1--59,
  2019{\natexlab{a}}.

\bibitem[Cotter et~al.(2019{\natexlab{b}})Cotter, Jiang, and
  Sridharan]{cotter2019two}
Andrew Cotter, Heinrich Jiang, and Karthik Sridharan.
\newblock Two-player games for efficient non-convex constrained optimization.
\newblock In \emph{Algorithmic Learning Theory}, pages 300--332. PMLR,
  2019{\natexlab{b}}.

\bibitem[Cui et~al.(2019)Cui, Jia, Lin, Song, and Belongie]{cui2019class}
Yin Cui, Menglin Jia, Tsung-Yi Lin, Yang Song, and Serge Belongie.
\newblock Class-balanced loss based on effective number of samples.
\newblock In \emph{Proceedings of the IEEE/CVF conference on Computer Vision
  and Pattern Recognition, {CVPR}}, pages 9268--9277, 2019.

\bibitem[Devlin et~al.(2019)Devlin, Chang, Lee, and Toutanova]{devlin2018bert}
Jacob Devlin, Ming{-}Wei Chang, Kenton Lee, and Kristina Toutanova.
\newblock {BERT:} pre-training of deep bidirectional transformers for language
  understanding.
\newblock In \emph{Proceedings of the 2019 Conference of the North American
  Chapter of the Association for Computational Linguistics: Human Language
  Technologies, {NAACL-HLT}}, pages 4171--4186. Association for Computational
  Linguistics, 2019.

\bibitem[Dong et~al.(2017)Dong, Gong, and Zhu]{dong2017class}
Qi~Dong, Shaogang Gong, and Xiatian Zhu.
\newblock Class rectification hard mining for imbalanced deep learning.
\newblock In \emph{Proceedings of the IEEE International Conference on Computer
  Vision}, pages 1851--1860, 2017.

\bibitem[Esteva et~al.(2019)Esteva, Robicquet, Ramsundar, Kuleshov, DePristo,
  Chou, Cui, Corrado, Thrun, and Dean]{esteva2019guide}
Andre Esteva, Alexandre Robicquet, Bharath Ramsundar, Volodymyr Kuleshov, Mark
  DePristo, Katherine Chou, Claire Cui, Greg Corrado, Sebastian Thrun, and Jeff
  Dean.
\newblock A guide to deep learning in healthcare.
\newblock \emph{Nature Medicine}, 25\penalty0 (1):\penalty0 24--29, 2019.

\bibitem[Fallah et~al.(2021)Fallah, Mokhtari, and
  Ozdaglar]{fallah2021generalization}
Alireza Fallah, Aryan Mokhtari, and Asuman Ozdaglar.
\newblock Generalization of model-agnostic meta-learning algorithms: Recurring
  and unseen tasks.
\newblock \emph{arXiv preprint arXiv:2102.03832}, 2021.

\bibitem[Fan et~al.(2018)Fan, Tian, Qin, Li, and Liu]{fan2018learning}
Yang Fan, Fei Tian, Tao Qin, Xiang-Yang Li, and Tie-Yan Liu.
\newblock Learning to teach.
\newblock In \emph{International Conference on Learning Representations}, 2018.

\bibitem[Finn et~al.(2017)Finn, Abbeel, and Levine]{finn2017model}
Chelsea Finn, Pieter Abbeel, and Sergey Levine.
\newblock Model-agnostic meta-learning for fast adaptation of deep networks.
\newblock In \emph{International Conference on Machine Learning}, pages
  1126--1135. PMLR, 2017.

\bibitem[Hardt et~al.(2016)Hardt, Price, and Srebro]{hardt2016equality}
Moritz Hardt, Eric Price, and Nati Srebro.
\newblock Equality of opportunity in supervised learning.
\newblock In \emph{Advances in Neural Information Processing Systems 29}, pages
  3315--3323, 2016.

\bibitem[He et~al.(2016)He, Zhang, Ren, and Sun]{he2016deep}
Kaiming He, Xiangyu Zhang, Shaoqing Ren, and Jian Sun.
\newblock Deep residual learning for image recognition.
\newblock In \emph{Proceedings of the IEEE conference on computer vision and
  pattern recognition}, pages 770--778, 2016.

\bibitem[Hinton et~al.(2012)Hinton, Deng, Yu, Dahl, Mohamed, Jaitly, Senior,
  Vanhoucke, Nguyen, Sainath, et~al.]{hinton2012deep}
Geoffrey Hinton, Li~Deng, Dong Yu, George~E Dahl, Abdel-rahman Mohamed, Navdeep
  Jaitly, Andrew Senior, Vincent Vanhoucke, Patrick Nguyen, Tara~N Sainath,
  et~al.
\newblock Deep neural networks for acoustic modeling in speech recognition: The
  shared views of four research groups.
\newblock \emph{IEEE Signal processing magazine}, 29\penalty0 (6):\penalty0
  82--97, 2012.

\bibitem[Jiang and Nachum(2020)]{jiang2020identifying}
Heinrich Jiang and Ofir Nachum.
\newblock Identifying and correcting label bias in machine learning.
\newblock In \emph{The 23rd International Conference on Artificial Intelligence
  and Statistics, {AISTATS}}, pages 702--712. PMLR, 2020.

\bibitem[Jiang et~al.(2018)Jiang, Zhou, Leung, Li, and
  Fei-Fei]{jiang2018mentornet}
Lu~Jiang, Zhengyuan Zhou, Thomas Leung, Li-Jia Li, and Li~Fei-Fei.
\newblock Mentornet: Learning data-driven curriculum for very deep neural
  networks on corrupted labels.
\newblock In \emph{International Conference on Machine Learning}, pages
  2304--2313. PMLR, 2018.

\bibitem[Kahn and Marshall(1953)]{kahn1953methods}
Herman Kahn and Andy~W Marshall.
\newblock Methods of reducing sample size in monte carlo computations.
\newblock \emph{Journal of the Operations Research Society of America},
  1\penalty0 (5):\penalty0 263--278, 1953.

\bibitem[Krizhevsky et~al.(2012)Krizhevsky, Sutskever, and
  Hinton]{krizhevsky2012imagenet}
Alex Krizhevsky, Ilya Sutskever, and Geoffrey~E. Hinton.
\newblock Imagenet classification with deep convolutional neural networks.
\newblock In \emph{Advances in Neural Information Processing Systems}, pages
  1106--1114, 2012.

\bibitem[Kumar et~al.(2010)Kumar, Packer, and Koller]{kumar2010self}
M~Kumar, Benjamin Packer, and Daphne Koller.
\newblock Self-paced learning for latent variable models.
\newblock \emph{Advances in neural information processing systems},
  23:\penalty0 1189--1197, 2010.

\bibitem[Lahoti et~al.(2020)Lahoti, Beutel, Chen, Lee, Prost, Thain, Wang, and
  Chi]{lahoti2020fairness}
Preethi Lahoti, Alex Beutel, Jilin Chen, Kang Lee, Flavien Prost, Nithum Thain,
  Xuezhi Wang, and Ed~H Chi.
\newblock Fairness without demographics through adversarially reweighted
  learning.
\newblock In \emph{Advances in Neural Information Processing Systems 33}, 2020.

\bibitem[Lake et~al.(2015)Lake, Salakhutdinov, and Tenenbaum]{lake2015human}
Brenden~M Lake, Ruslan Salakhutdinov, and Joshua~B Tenenbaum.
\newblock Human-level concept learning through probabilistic program induction.
\newblock \emph{Science}, 350\penalty0 (6266):\penalty0 1332--1338, 2015.

\bibitem[Lin et~al.(2017)Lin, Goyal, Girshick, He, and
  Doll{\'a}r]{lin2017focal}
Tsung-Yi Lin, Priya Goyal, Ross Girshick, Kaiming He, and Piotr Doll{\'a}r.
\newblock Focal loss for dense object detection.
\newblock In \emph{Proceedings of the {IEEE} International Conference on
  Computer Vision, {ICCV}}, pages 2980--2988, 2017.

\bibitem[Malisiewicz et~al.(2011)Malisiewicz, Gupta, and
  Efros]{malisiewicz2011ensemble}
Tomasz Malisiewicz, Abhinav Gupta, and Alexei~A Efros.
\newblock Ensemble of exemplar-svms for object detection and beyond.
\newblock In \emph{2011 International conference on computer vision}, pages
  89--96. IEEE, 2011.

\bibitem[Petrovi{\'c} et~al.(2020)Petrovi{\'c}, Nikoli{\'c}, Radovanovi{\'c},
  Deliba{\v{s}}i{\'c}, and Jovanovi{\'c}]{petrovic2020fair}
Andrija Petrovi{\'c}, Mladen Nikoli{\'c}, Sandro Radovanovi{\'c}, Boris
  Deliba{\v{s}}i{\'c}, and Milo{\v{s}} Jovanovi{\'c}.
\newblock Fair: Fair adversarial instance re-weighting.
\newblock \emph{arXiv preprint arXiv:2011.07495}, 2020.

\bibitem[Platt and Barr(1987)]{platt1987constrained}
John~C Platt and Alan~H Barr.
\newblock Constrained differential optimization.
\newblock In \emph{Proceedings of the 1987 International Conference on Neural
  Information Processing Systems}, pages 612--621, 1987.

\bibitem[Ren et~al.(2018)Ren, Zeng, Yang, and Urtasun]{ren2018learning}
Mengye Ren, Wenyuan Zeng, Bin Yang, and Raquel Urtasun.
\newblock Learning to reweight examples for robust deep learning.
\newblock In \emph{Proceedings of the 35th International Conference on Machine
  Learning, {ICML}}, pages 4334--4343. PMLR, 2018.

\bibitem[Roh et~al.(2021)Roh, Lee, Whang, and Suh]{roh2021sample}
Yuji Roh, Kangwook Lee, Steven~Euijong Whang, and Changho Suh.
\newblock Sample selection for fair and robust training.
\newblock In \emph{Thirty-Fifth Conference on Neural Information Processing
  Systems}, 2021.

\bibitem[Saxena et~al.(2019)Saxena, Tuzel, and DeCoste]{saxena2019data}
Shreyas Saxena, Oncel Tuzel, and Dennis DeCoste.
\newblock Data parameters: {A} new family of parameters for learning a
  differentiable curriculum.
\newblock In \emph{Advances in Neural Information Processing Systems}, pages
  11093--11103, 2019.

\bibitem[Seto et~al.(2021)Seto, Wells, and Zhang]{seto2021halo}
Skyler Seto, Martin~T Wells, and Wenyu Zhang.
\newblock Halo: Learning to prune neural networks with shrinkage.
\newblock In \emph{Proceedings of the 2021 SIAM International Conference on
  Data Mining (SDM)}, pages 558--566. SIAM, 2021.

\bibitem[Shu et~al.(2019)Shu, Xie, Yi, Zhao, Zhou, Xu, and Meng]{shu2019meta}
Jun Shu, Qi~Xie, Lixuan Yi, Qian Zhao, Sanping Zhou, Zongben Xu, and Deyu Meng.
\newblock Meta-weight-net: Learning an explicit mapping for sample weighting.
\newblock \emph{Advances in Neural Information Processing Systems},
  32:\penalty0 1919--1930, 2019.

\bibitem[Thrun and Pratt(2012)]{thrun2012learning}
Sebastian Thrun and Lorien Pratt.
\newblock \emph{Learning to learn}.
\newblock Springer Science \& Business Media, 2012.

\bibitem[Toneva et~al.(2018)Toneva, Sordoni, des Combes, Trischler, Bengio, and
  Gordon]{toneva2018empirical}
Mariya Toneva, Alessandro Sordoni, Remi~Tachet des Combes, Adam Trischler,
  Yoshua Bengio, and Geoffrey~J Gordon.
\newblock An empirical study of example forgetting during deep neural network
  learning.
\newblock In \emph{International Conference on Learning Representations}, 2018.

\bibitem[Vyas et~al.(2020)Vyas, Saxena, and Voice]{vyas2020learning}
Nidhi Vyas, Shreyas Saxena, and Thomas Voice.
\newblock Learning soft labels via meta learning.
\newblock \emph{arXiv preprint arXiv:2009.09496}, 2020.

\bibitem[Wu et~al.(2018)Wu, Tian, Xia, Fan, Qin, Lai, and Liu]{wu2018learning}
Lijun Wu, Fei Tian, Yingce Xia, Yang Fan, Tao Qin, Jian-Huang Lai, and Tie-Yan
  Liu.
\newblock Learning to teach with dynamic loss functions.
\newblock In \emph{NeurIPS}, 2018.

\bibitem[Zadrozny(2004)]{zadrozny2004learning}
Bianca Zadrozny.
\newblock Learning and evaluating classifiers under sample selection bias.
\newblock In \emph{Proceedings of the twenty-first international conference on
  Machine learning}, page 114, 2004.

\bibitem[Zhao et~al.(2019)Zhao, Fard, Narasimhan, and Gupta]{zhao2019metric}
Sen Zhao, Mahdi~Milani Fard, Harikrishna Narasimhan, and Maya~R. Gupta.
\newblock Metric-optimized example weights.
\newblock In \emph{Proceedings of the 36th International Conference on Machine
  Learning, {ICML}}, volume~97 of \emph{Proceedings of Machine Learning
  Research}, pages 7533--7542. {PMLR}, 2019.

\end{thebibliography}

\appendix
\onecolumn
\icmltitle{FORML: Learning to Reweight Data for Fairness - Supplementary Material}
\section{FORML Hyperparameters: Sensitivity and Variations}

\subsection{FORML Hyperparameters}

\noindent \textbf{CIFAR:} For CIFAR-10/100, we train a ResNet-18 architecture \citep{he2016deep} using SGD with Nesterov momentum $\gamma=0.9$ and batch size of $128$.  We set the initial learning rate to $\eta_c=0.1$ and decay by $0.1$ at the $60$\textsuperscript{th}, $120$\textsuperscript{th}, and $160$\textsuperscript{th} epochs and train for a total of $200$ epochs. We also penalize the model parameters with weight decay at $5e-4$. For both CIFAR-10/100, we use standard data augmentations (normalization, random horizontal flip, translation by up to 4 pixels).  To learn the training sample weights in FORML, we use SGD with the same learning rate as the model parameter learning rate and minimize the max loss discrepancy.  For MOEW on CIFAR-10 we used twenty batches for learning the weight function parameters and fifty batches on CIFAR-100.  Both MOEW and FORML use a held-out validation set of 5000 samples for learning weights and maintains a minimum of 3 samples per class per batch.

\noindent \textbf{Celeb-A:}  For Celeb-A, we train a ResNet-18 architecture \citep{he2016deep} to predict the binary attribute ``Attractive'' given images of faces. The sensitive attribute is gender denoted by ``Male'' in the dataset. The model is trained using Adam optimizer and batch size of $256$.  We set the initial learning rate to $\eta_c=0.001$ and train for a total of $10$ epochs. For pre-processing the images, we center-crop and resize to size $128\times 128$ and normalize.  To learn the training sample weights in FORML, we use SGD with the same learning rate as the model parameter learning rate and minimize the mean loss discrepancy.  The validation set maintains a minimum of 20 samples per class per batch for FORML.  For the proportion weighting, we take the number of samples according to the sensitive attribute because the number of samples in each group for the target attribute ``Attractive'' are nearly identical at $51.4\%$ in the training set and we seek to reduce fairness violations in the sensitive attribute ``Male'' in the dataset.  For this dataset, we use the pre-defined train, val, test split and compare models at the epoch with the highest validation accuracy during training. Metrics are reported over the test set

\noindent \textbf{MNIST}: For the corrupt-label MNIST experiment, a two hidden layer network with ReLU activations is used. For all methods, we select $5$k samples randomly from the training set for validation and use early stopping to prevent overfitting to the corrupt labels. We use the same training hyperparameters as in \citep{jiang2020identifying}.

\subsection{FORML Variations}

We conduct ablation studies on the Celeb-A dataset to determine the impact of different hyperparameters and settings in FORML.  In the main paper, we use the best performing hyperparameters that also reduce the number of hyperparameters needed to tune (i.e. maintaining the same learning rate, and only reported results with two meta-loss functions, although there are many others that are compatible with FORML).

\noindent \textbf{Reverse and Adversarial comparisons:} Our algorithm shares many similarities with the Lagrange multipliers and basic differentiable multipliers method (BDMM) algorithms \citep{platt1987constrained}, which have been applied for constrained optimization problems and in recent work for fairness \citep{cotter2019optimization, jiang2020identifying}.  In the BDMM algorithm, it is argued that standard gradient descent for the constraint in Lagrange multipliers does not work and instead  the reverse sign should be applied \citep{platt1987constrained}.  We experiment with both a reverse sign in the update for the model parameters in the first update, as well as for the weights of the model and report results in \autoref{celeba-reverse} and \autoref{celeba-adversarial}.  Results indicate that FORML performs better when using the reverse update on the model parameters, but the reverse update on the sample weights has little effect.  While results are slightly higher for the test accuracy, TPRD on the validation set was on average $1\%$ lower for the reverse update on the sample weights. Thus our algorithm in the main paper uses a reverse update for the model parameters, and not for the sample weights indicating some differences from the BDMM algorithm.

\begin{table}[h]
  \centering
  \begin{tabular}{lccc}
    \toprule
    Method                    & TPRD (\%)  $\downarrow$           & maxFNR (\%) $\downarrow$          & Accuracy (\%)    $\uparrow$      \\
    \midrule
    No Reverse & $22.80 \pm 3.88$ & $37.35 \pm 7.55$ & $\bm{79.99 \pm 1.11}$\\
    Reverse & $\bm{20.41 \pm 3.34}$ & $\bm{33.89 \pm 8.34}$ & $79.48 \pm 0.82$ \\

    \bottomrule
  \end{tabular}
    \caption{Testing the model with a reverse update for the model parameters by comparing the True Positive Rate Disparity (TPRD), Maximum False Negative Rate (maxFNR), and test accuracy on Celeb-A. The means and standard errors are from 5 runs. $\downarrow$ denotes lower values are better. $\uparrow$ denotes higher values are better.}
    \label{celeba-reverse}
\end{table}

\begin{table}[ht]
  \centering
  \begin{tabular}{lccc}
    \toprule
    Method                    & TPRD (\%)  $\downarrow$           & maxFNR (\%) $\downarrow$          & Accuracy (\%)    $\uparrow$      \\
    \midrule
    No Reverse & $\bm{20.41 \pm 3.34}$ & $33.89 \pm 8.34$ & $79.48 \pm 0.82$\\
    Reverse & $20.45 \pm 2.44$ & $\bm{33.32 \pm 6.4}8$ & $\bm{80.37 \pm 0.57}$ \\
    \bottomrule
  \end{tabular}
    \caption{Testing the model with a reverse update for the model parameters by comparing the True Positive Rate Disparity (TPRD), Maximum False Negative Rate (maxFNR), and test accuracy on Celeb-A. The means and standard errors are from 5 runs. $\downarrow$ denotes lower values are better. $\uparrow$ denotes higher values are better.}
    \label{celeba-adversarial}
\end{table}

\noindent \textbf{Meta Loss Variations:} In Section~3, the meta-loss optimized over the validation set is either the MaxLossD or the MeanLossD which generally perform similarly.  We demonstrate the advantage of using a loss oriented for fairness in Table~\ref{celeba-metaloss} by comparing with the cross entropy loss.  We find that using MaxLossD and MeanLossD lower the fairness violation, but  marginally lowers accuracy; whereas using cross entropy increases accuracy slightly but does not lower the fairness violation with respect to baseline performance from Table~3.

\begin{table}[ht]
  \centering
  \begin{tabular}{lccc}
    \toprule
    Method                    & TPRD (\%)  $\downarrow$           & maxFNR (\%) $\downarrow$          & Accuracy (\%)    $\uparrow$      \\
    \midrule
    MaxLossD & $20.85 \pm 2.88$ & $37.66 \pm 7.96$ & $79.41 \pm 0.59$\\
    MeanLossD & $\bm{20.41 \pm 3.34}$ & $\bm{33.89 \pm 8.34}$ & $79.48 \pm 0.82$\\
    Cross Entropy & $24.70 \pm 2.74$ & $38.87 \pm 4.76$ & $\bm{81.08 \pm 0.48}$\\
    \bottomrule
  \end{tabular}
    \caption{Testing the model with cross entropy as the meta-loss by comparing the True Positive Rate Disparity (TPRD), Maximum False Negative Rate (maxFNR), and test accuracy on Celeb-A. The means and standard errors are from 5 runs. $\downarrow$ denotes lower values are better. $\uparrow$ denotes higher values are better.}
    \label{celeba-metaloss}
\end{table}

\noindent \textbf{Sample Weight Variations:} In Section~3, the weights are directly multiplied to balance the loss.  However, other variations of weights could be applied to balance the loss such as weights on the logits, or a different function of the weight parameters to enforce different behavior.  Here, we explore a reweighting based on the logits, and a weighting based on the $L_1$ penalty that encourages sparsity of the weights by forcing weights that are too large to zero; we denote this weighting as the inverse square $L_1$ \citep{seto2021halo}.  Both are compared with the baseline of directly multiplying the learned sample weights with the sample loss, and results reported in \autoref{celeba-weighttype} indicate that the baseline reweighting performs best across all metrics.

\begin{table}[ht]
  \centering
  \begin{tabular}{lccc}
    \toprule
    Method                    & TPRD (\%)  $\downarrow$           & maxFNR (\%) $\downarrow$          & Accuracy (\%)    $\uparrow$      \\
    \midrule
    Baseline & $\bm{18.91 \pm 1.16}$ & $\bm{28.57 \pm 2.27}$ & $\bm{80.91 \pm 0.30}$\\
    Logits & $25.32 \pm 2.36$ & $43.81 \pm 5.96$ & $80.65 \pm 0.58$\\
    Inverse Sq. $L_1$ & $24.52 \pm 3.67$ & $44.08 \pm 9.50$ & $78.90 \pm 1.02$	\\
    \bottomrule
  \end{tabular}
    \caption{Testing the model with different methods for applying weights by comparing the True Positive Rate Disparity (TPRD), Maximum False Negative Rate (maxFNR), and test accuracy on Celeb-A. The means and standard errors are from 5 runs. $\downarrow$ denotes lower values are better. $\uparrow$ denotes higher values are better.}
    \label{celeba-weighttype}
\end{table}

\section{Alternative CelebA Tasks}

In addition to the task of predicting the ``Attractive'' attribute with gender (denoted ``Male'') as the sensitive attribute, we also experiment with different tasks and sensitive attributes to test robustness of FORML to different tasks.  We change the sensitive attribute to the ``Pale Skin'' attribute (color), and the prediction task to the ``Young'' attribute.  Accuracy on this task is higher than for the ``Attractive'' attribute, and as such the fairness violation is lower.  Nonetheless,  we report in Table~\ref{celeba-alternate} that FORML is able to reduce fairness violations over uniform weighting.

\begin{table}[ht]
  \centering
  \begin{tabular}{lccc}
    \toprule
    Method                    & TPRD (\%)  $\downarrow$           & maxFNR (\%) $\downarrow$          & Accuracy (\%)    $\uparrow$      \\
    \midrule
    Uniform & $2.64 \pm 0.46$& $5.36 \pm 0.25$ & $\bm{87.62 \pm 0.09}$\\
    FORML & $\bm{2.51 \pm 0.43}$ & $\bm{4.64 \pm 0.84}$ & $87.07 \pm 0.22$\\
    \bottomrule
  \end{tabular}
    \caption{Evaluating FORML for predicting ``Young'' subject to the sensitive attribute color (``Pale Skin'') according to the metrics: True Positive Rate Disparity (TPRD), Maximum False Negative Rate (maxFNR), and test accuracy on Celeb-A. The means and standard errors are from 5 runs. $\downarrow$ denotes lower values are better. $\uparrow$ denotes higher values are better.}
    \label{celeba-alternate}
\end{table}

\section{More on the Normalized Loss for Reducing Overfitting to Corrupt Labels}

\label{expanded-mnist-corrupt}

In Section~4.1, FORML is tested in the robustness setting with incorrect labels.  We further explore how training with FORML improves performance by examining the loss for each sample over epochs during training.  The objective of this analysis is to investigate how the loss differs for the model trained with FORML, and how this leads to improvement on the test set.  

We compute an exponential moving average (EMA) on the loss and compare the average EMA loss for samples with label ``2'' to samples with another label.  As expected, since many of the samples have an incorrect label ``2'', the loss for samples with label ``2'' is higher in both a model trained with uniform weights and the model trained with FORML.  For a model with uniform weights, the ratio of the losses for label ``2'' samples to other samples is $2.9$.  In comparison, the model trained with FORML has a ratio of $2.61$, a relative decrease of $10\%$. As the accuracy for both models in predicting samples with label ``2'' are relatively similar, we can conclude it is likely that the weights learned by FORML are reducing the model's ability to overfit to the incorrect labels leading to better performance on the test set.

\end{document}